\documentclass[letterpaper,10pt,conference]{ieeeconf}
\IEEEoverridecommandlockouts
\usepackage{amsmath,amsfonts}
\usepackage{algorithm}
\usepackage{algpseudocode}
\usepackage{array}
\usepackage[caption=false,font=normalsize,labelfont=sf,textfont=sf]{subfig}
\usepackage{textcomp}
\usepackage{stfloats}
\usepackage{url}
\usepackage{verbatim}
\usepackage{graphicx}
\usepackage{cite}
\usepackage{orcidlink}
\usepackage{multirow}
\PassOptionsToPackage{bookmarks=false}{hyperref}
\hypersetup{hidelinks}

\newif\ifshownotes
\shownotestrue

\begin{document}

\title{RAFAIL: Relationship-Aware Failure Detection for Robotic Manipulation}

\author{Loris~Schneider$^*$, Edgar~Welte$^*$, Rania Rayyes
\thanks{$^*$Equal contribution. \newline Institute for Material Handling and Logistics (IFL), Karlsruhe Institute of Technology, 76131, Germany. Email: \texttt{loris.schneider@kit.edu}. \newline This work is supported by the German Federal Ministry of Research, Technology, and Space (BMFTR) under the Robotics Institute Germany (RIG), the DFG SFB-1574-471687386 project, and the Ministry of Science, Research and Arts of the Federal State of Baden-Württemberg within the InnovationCampus Future Mobility.}}

\maketitle

\vspace{1em}
\setlength{\textfloatsep}{-1pt}

\begin{abstract}

Detecting failures during execution is essential for reliable robotic manipulation. Vision-language models (VLMs) can assess task outcomes semantically but add runtime computation, whereas out-of-distribution (OOD) detectors may respond to harmless scene variations rather than failure-relevant deviations. We introduce RAFAIL, a framework for detecting execution failures during robotic manipulation. RAFAIL identifies failures by detecting anomalies in task-relevant relationships between entities, such as a gripper and an object or an object and its target. By focusing OOD detection on relevant parts of the observation, RAFAIL reduces sensitivity to task-irrelevant scene variation. Offline, a VLM annotates successful demonstrations with task progress and relationship importance, which are used to learn point-cloud-based relationship representations without relying on policy-internal features. At runtime, relationship-specific OOD detectors evaluate these representations while relationship importance and task progress are predicted without VLM inference.
RAFAIL requires no failure data and achieves 73.4\% balanced accuracy across three real-world robotic manipulation tasks, outperforming the strongest evaluated OOD- and uncertainty-based baselines.

\end{abstract}

\section{Introduction}
\label{sec:introduction}

Robot learning has enabled increasingly complex manipulation tasks~\cite{wolfDiffusionModelsRobotic2025, welteInteractiveImitationLearning2025}, yet learned policies remain prone to failures during real-world execution due to uncertainty in the environment, perception, and unexpected physical interaction. Detecting such failures is essential for reliable autonomous operation, allowing execution to be stopped or recovery to be initiated before failures cause damage~\cite{roemer2025failureprediction, lin2025failsafe, gu2025safe}.
The central challenge is not merely detecting that an execution differs from previously successful behavior, but determining whether that deviation affects task success.

Detectors leveraging visual language models (VLMs) use visual observations and task descriptions to assess whether execution satisfies a goal~\cite{duan2024aha,lin2025failsafe,grislain2026ifailsense}. Although they can recognize semantic failures, continuous VLM inference adds substantial runtime cost~\cite{gu2025safe,rolland2026failureidentification}. Out-of-distribution (OOD) and uncertainty-based detectors can instead be trained on successful executions and evaluated efficiently during deployment~\cite{xu2025faildetect,roemer2025failureprediction}. However, their signals indicate deviation or uncertainty rather than failure: a harmless change in the scene can trigger an alert, while a subtle but consequential interaction error may be missed~\cite{rolland2026failureidentification}. Many such detectors additionally rely on policy-internal features that may not be easily available across policy architectures~\cite{gu2025safe}.

We address these challenges by focusing failure detection on task-relevant relationships between entities rather than global observations or policy representations. In manipulation, task success often depends on how these relationships evolve, such as between a gripper and an object or an object and its target. We therefore introduce \underline{r}elationship-\underline{a}ware \underline{fail}ure detection (RAFAIL), a framework for robustly detecting execution failures during robotic manipulation. RAFAIL represents individual relationships from point-cloud observations and applies task-progress-conditioned and importance-gated OOD detection to each. A VLM provides semantic supervision only offline by annotating successful demonstrations with task progress and relationship importance, while runtime detection requires neither VLM inference nor access to policy-internal features. Relationship-specific OOD scores additionally indicate which interaction deviates from successful execution. Our contributions are:

\begin{figure}
    \centering
    \includegraphics[width=0.95\linewidth]{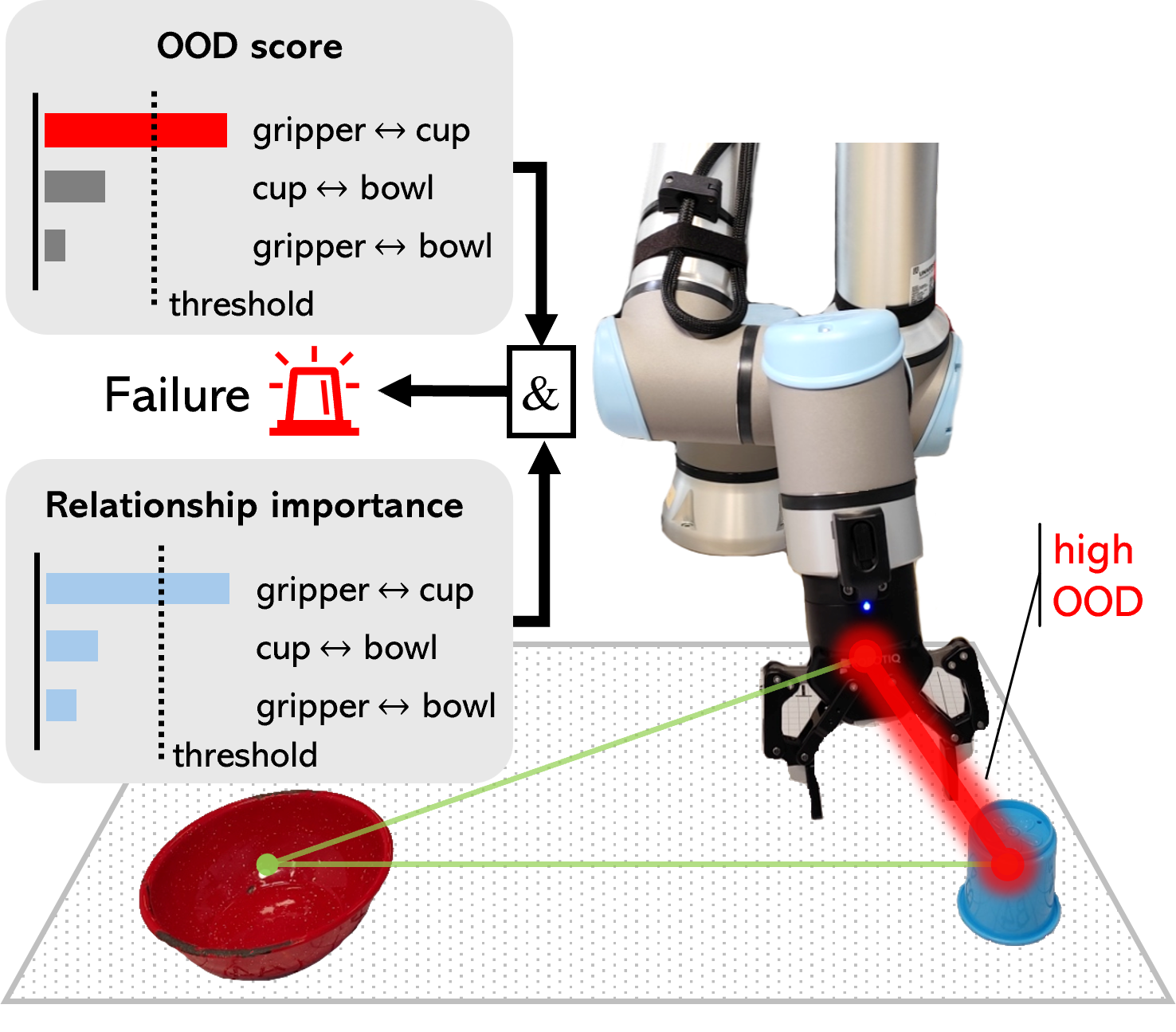}
    \caption{RAFAIL detects failures by monitoring task-relevant relationships. Relationship importance, estimated from successful demonstrations, gates the contribution of each relationship at runtime. A failure is triggered when an important relationship exhibits a high OOD score, while deviations in less relevant relations are suppressed.}
    \label{fig:teaser}
\end{figure}

\begin{itemize}
    \item \textbf{A relationship-aware framework for failure detection} that decomposes global observations into dedicated representations of individual task-relevant relationships.
    \item \textbf{A relationship importance and task progress estimation approach for failure detection} via distillation of semantic task understanding from a VLM, enabling relationship importance and task progress estimates without online VLM inference.
     \item \textbf{A task-progress-conditioned relationship OOD detector} that localizes which interaction is anomalous, integrates task progress and relationship importance, and
is independent of policy-internal latent representations.

\end{itemize}
RAFAIL is evaluated on three real-world robotic manipulation tasks, achieving a balanced accuracy of 73.4\% versus 65.2\% for the best evaluated baseline.

\section{Related Work}
\label{sec:related_work}

\subsection{Foundation Models for Failure Detection}
The semantic understanding of foundation models, especially VLMs, makes them attractive for application in failure detection as they can derive task understanding and task success conditions.
AHA~\cite{duan2024aha} finetunes a VLM on synthetic manipulation failures for detection and explanation, while FailSafe~\cite{lin2025failsafe} additionally learns recovery behavior. I-FailSense~\cite{grislain2026ifailsense} uses VLM features for semantic failure classification.
Their framework outperforms larger non-finetuned VLMs, as well as AHA \cite{duan2024aha}, in failure detection tasks.\\
Those works show that adapted foundation models, especially VLMs, work well for failure detection and reasoning, especially for the detection of semantic failures. But continuously running VLM inference alongside a policy creates a large computational overhead \cite{gu2025safe, rolland2026failureidentification}. This is additional to the high computational demands of finetuning VLMs on large failure datasets. In contrast, RAFAIL does not rely on a VLM for online inference.

\subsection{Uncertainty- and OOD-based Failure Detection}
A second branch of failure detection methods is based on the intuition that policies are likely to fail in OOD situations and therefore aims to detect such situations rather than directly detecting failure. The general approach was described in \cite{xu2025faildetect} and called Fail-Detect. It consists of extracting signals from policy inputs and outputs during successful task demonstrations and training OOD-detectors on them. Then, they calculate Conformal Prediction (CP) \cite{angelopoulos2023conformalprediction} thresholds for those OOD scores. They evaluate the performance of diverse OOD-detectors for robot manipulation tasks. In real-world experiments, they identify their own logpZO \cite{xu2025faildetect} and Random Network Distillation (RND)\cite{burda2019rnd} as the best performing learned detectors but also compare to non-learned methods like PCA-kmeans \cite{liu2024pcakmeans, xu2025faildetect}.
While \cite{xu2025faildetect} evaluates OOD-detectors on observations, predicted actions, or both, \cite{roemer2025failureprediction} propose to also estimate the uncertainty of the action prediction. Their framework named FIPER uses RND as an OOD-detector on the observations. Additionally, FIPER samples batches of action chunks from the policy at each timestep and estimates their entropy as an uncertainty measure which they call Action Chunk Entropy (ACE). A failure label is only produced if both the RND OOD-detector and the uncertainty estimator exceed their respective CP thresholds. They show consistently high failure detection accuracies with this combined approach, outperforming pure OOD-detection or uncertainty estimation on most tasks.

Uncertainty- and OOD-based failure detection approaches are generally more computationally efficient and can run alongside a policy with little overhead. Crucially, they do not require additional failure data, using only successful trajectories. Finally, they allow failures to be predicted before they occur.
On the other hand, since those approaches detect deviations from successful inputs and/or outputs instead of actual failures, they are usually not able to distinguish failures from harmless deviations and might struggle to detect subtle failures \cite{rolland2026failureidentification}.

The authors of \cite{rolland2026failureidentification} correctly identified this limitation and proposed to deploy a VLM as an additional semantic filter. If an OOD-detector passes its CP threshold, a heatmap is created that marks the location of the deviation in an RGB image. This image is input into a VLM which decides if the deviation is harmless or a failure. Since they use an off-the-shelf, non-finetuned VLM and query it only in OOD situations, they do not require failure data and reduce runtime computation demand, but still require full VLM inference during runtime.

In contrast to prior OOD- and uncertainty-based approaches that operate on global observations, actions, or policy representations, RAFAIL performs OOD detection on task-relevant object relationships. Semantic information from a VLM is used only offline to identify and weight these relationships, while runtime detection remains lightweight and independent of policy-internal features.

\section{Method}
\label{sec:method}

\subsection{Problem Formulation}

\begin{figure}
    \centering
    \includegraphics[width=1\linewidth]{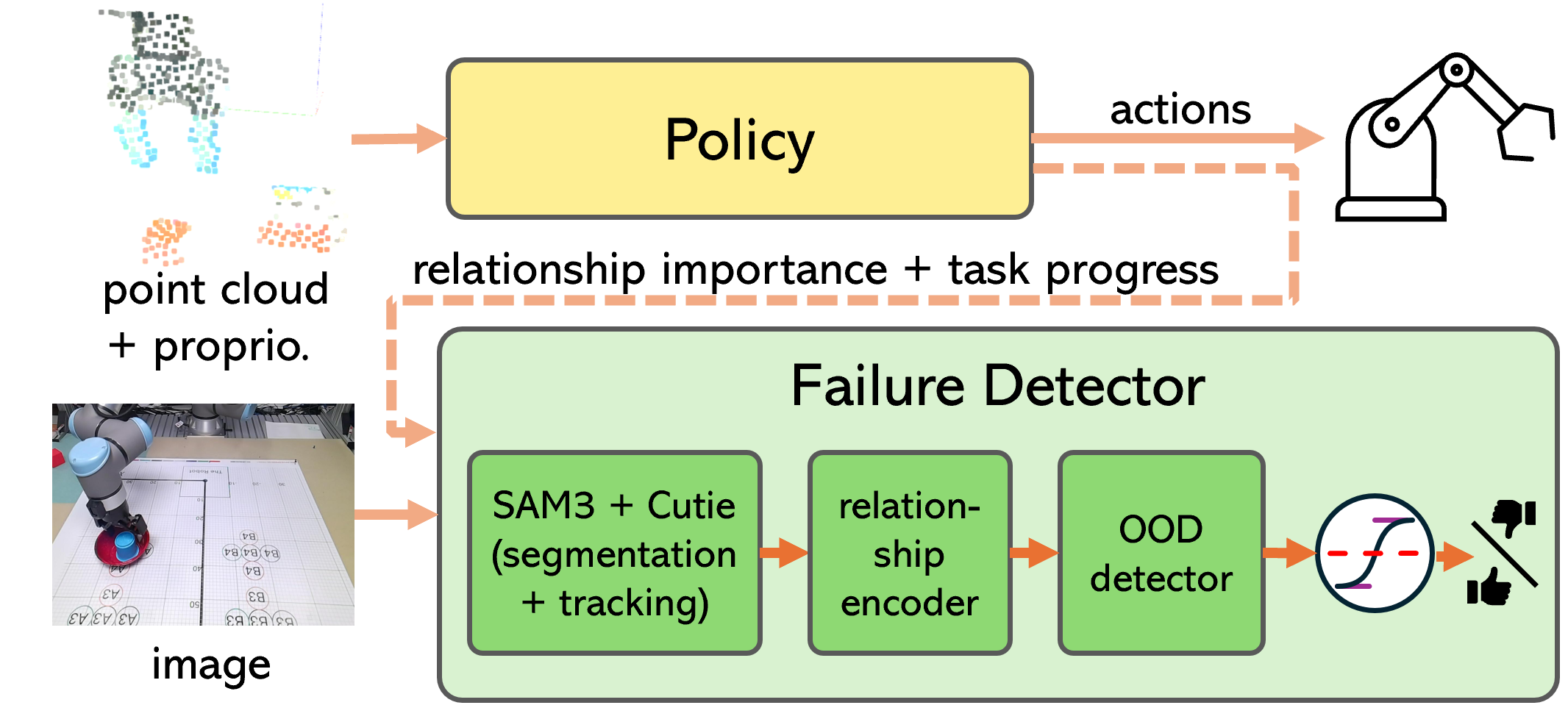}
    \caption{Overview of online deployment. The policy predicts actions based on point cloud and proprioceptive observations. It also predicts auxiliary scores for relationship importance and task progress. The failure detector module uses the image observation and the auxiliary scores to segment and score each relationship via an OOD detector.}
    \label{fig:online-overview}
\end{figure}

\paragraph{Imitation Learning Policy}
A 6-DoF robotic arm equipped with a parallel-jaw gripper operates over discrete control time steps \(t \in \{0,\ldots,T\}\). At each time step \(t\), the robot receives an observation
$$
\boldsymbol{o}_t = \left(\boldsymbol{o}_t^{\mathrm{pcd}},\,\boldsymbol{o}_t^{\mathrm{prio}}\right),
$$
consisting of a workspace point cloud \(\boldsymbol{o}_t^{\mathrm{pcd}} \in \mathbb{R}^{N^{\mathrm{pts}}\times 3}\) and the robot's proprioceptive state \(\boldsymbol{o}_t^{\mathrm{prio}} \in \mathrm{SE}(3)\times\{0,1\}\). The robot executes an end-effector command $ \boldsymbol{a}_t = \left(\boldsymbol{T}_t,\,g_t\right),$ where \(\boldsymbol{T}_t \in \mathrm{SE}(3)\) specifies the desired end-effector pose relative to the robot base, and \(g_t \in \{0,1\}\) denotes the binary gripper state (open or closed). The commanded pose is tracked by a low-level Cartesian impedance controller.
Given the most recent \(K\) observations, a policy \(\pi\) predicts a sequence of \(H\) future actions according to
$$
\hat{\boldsymbol{a}}_{t:t+H-1}
=
\pi\!\left(\boldsymbol{o}_{t-K+1:t}\right).
$$
In addition to action prediction, the policy exposes auxiliary estimates of task progress and object relationship importance, which the failure predictor uses. Figure~\ref{fig:online-overview} shows the pipeline overview during online deployment.

\paragraph{Failure Predictor}
We employ an independent failure predictor that estimates whether the current execution is likely to fail. At each time step \(t\), the predictor receives the current scene RGB and depth image \(\boldsymbol{o}_t^{\mathrm{img}}, \boldsymbol{o}_t^{\mathrm{depth}}\) together with auxiliary semantic information
$
\boldsymbol{c}_t = \left(\boldsymbol{r}_t,\,p_t\right),
$
where \(\boldsymbol{r}_t\) denotes the estimated importance of the relationships and \(p_t\) represents the current task progress. Based on these inputs, the failure predictor outputs a failure signal \(f_t^{\mathrm{fail}}\), indicating the possibility of an execution failure at time step \(t\).

\begin{figure*}
    \centering
    \includegraphics[width=1\linewidth]{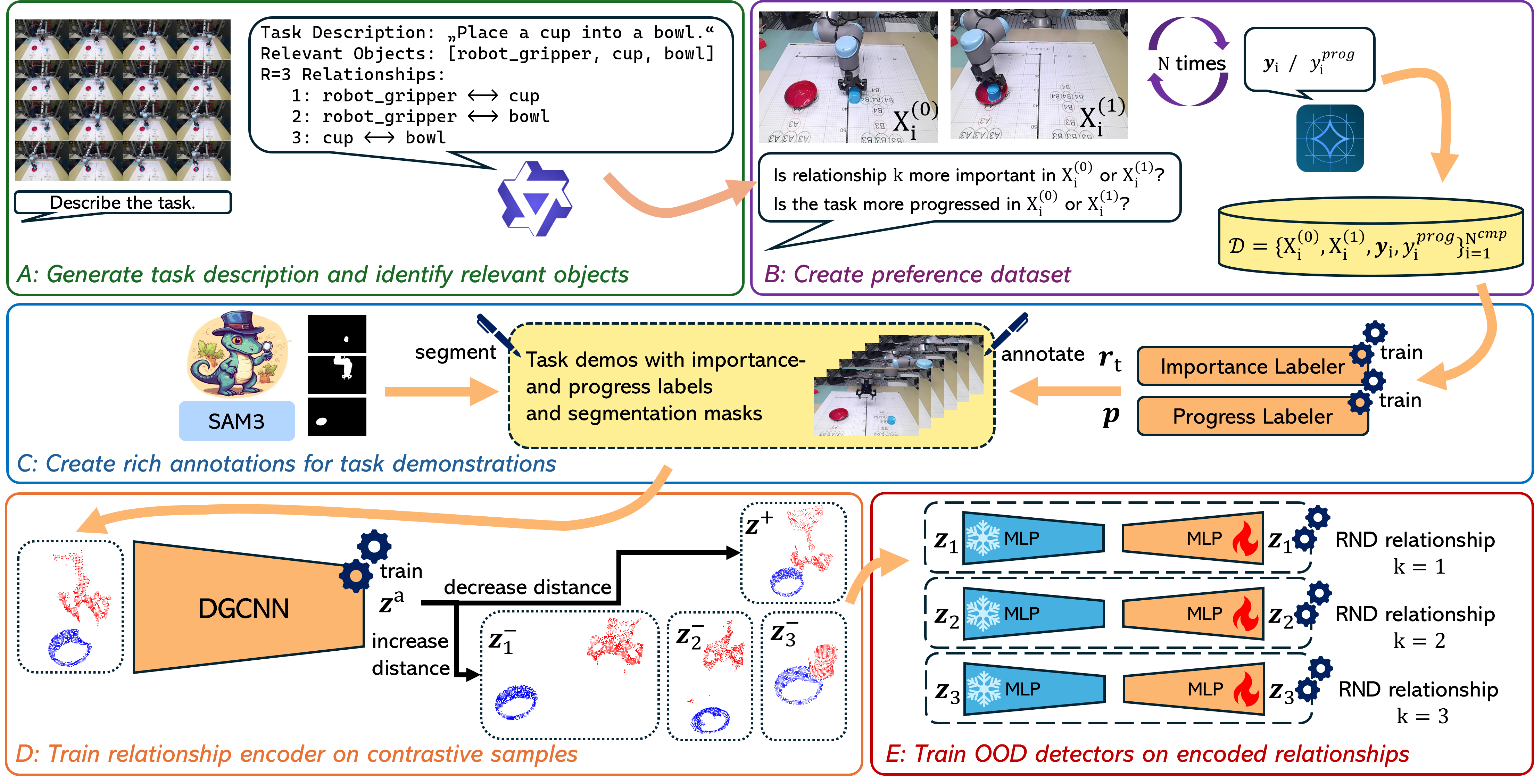}
    \caption{Overview of the offline training process.
    (A) Initially, a task is presented to a VLM as a sequence of images to extract a detailed task description, including relevant object relationships.
    (B) In the second stage, a preference dataset is created by labeling the importance of relationships and task progress using a smaller VLM.
    (C) The preference dataset is distilled into two labeling networks to generate a rich dataset for subsequent encoder and policy training.
    (D) The relationship encoder is trained to create an enriched, relationship-aware latent representation.
    (E) Out-of-distribution (OOD) detectors for each relationship are trained exclusively on successful demonstration data.}
    \label{fig:offline-overwiew}
\end{figure*}

\subsection{Method Overview}

Figure \ref{fig:offline-overwiew} summarizes the offline training pipeline.
During training, the VLM-generated pairwise preferences supervise the importance and progress labelers. These labelers are used only offline to annotate every successful demonstration. Their frame-wise outputs then serve as supervised targets for auxiliary heads attached to the manipulation policy. Consequently, neither the VLM nor the importance/progress labelers are executed at deployment; the policy directly predicts relationship importance $\hat r_t$ and task progress $\hat p_t$ jointly with its action output.

\subsection{Object-Relationship Representation}
We hypothesize that many task-level manipulation failures manifest as incorrect relationships between task-relevant entities. Entities refer to movable objects, stationary environment elements, and robot parts relevant to the current task. An object relationship describes the specific state of a pair of objects. We therefore focus only on object pairs to detect failures or abnormalities. We extract task-relevant objects from a scene and task description generated by a scene-understanding VLM. We provide a VLM \cite{qwen35} with frames from a task demonstration arranged in a single grid image and prompt it to describe the task, its start and goal states, and the task-relevant objects. We then refine the initial task description by providing grid images of different demonstrations of the same task, resulting in a more general description and alternative object descriptors. From the list of task-relevant objects, we create all unique object pairs and use those as the task-relevant object relationships.

\subsection{Relationship Importance Estimation}
The importance of an object relationship at a specific task stage must be inferred from semantic information while considering the task context. This abstract concept is not easily quantifiable. Therefore, we distill numeric relationship-importance labels from a VLM that can relate semantic information in an observation to the task context and goal. First, we provide a VLM \cite{gemmateam2026gemma4} with the task description along with two grid images, each showing a set of $L=4$ consecutive frames $X=(I_t, ... I_{t+L-1})$ with a random start frame $I_t$ and taken from a random task demonstration. We prompt the VLM to rank the importance of all object relationships between the two frame sequences given the task description and allowing for equal importance. Additionally, we also prompt the VLM to decide which of the two frame sequences corresponds to further task progress, also allowing for equal task progress. We collect a dataset
$\mathcal{D}=\{ X^{(0)}_i, X^{(1)}_i,\mathbf{y}_i, y_i^{\mathrm{prog}}\}^{N^{\mathrm{cmp}}}_{i=1},$
where $\mathbf{y_i}$ denotes the ranking of the relationship importance and $y^{\mathrm{prog}}_i$ denotes the ranking of the task progress with
\begin{equation*}
  y^{k}_i = \begin{cases}
      -1, \text{ if relationship $k$ is more important in $X^{(0)}_i$}\\
      0, \text{ if relationship $k$ is equally important in both}\\
      +1, \text{ if relationship $k$ is more important in  $X^{(1)}_i$}
  \end{cases}
\end{equation*}
and
\begin{equation*}
    y_i^{\mathrm{prog}} = \begin{cases}
        -1, \text{ if $X^{(1)}_i$ is chronologically before $X^{(0)}_i$}\\
        0, \text{ if both correspond to the same task progress}\\
        +1, \text{ if $X^{(0)}_i$ is chronologically before $X^{(1)}_i$}.
    \end{cases}
\end{equation*}
We use this dataset to train an importance labeler $\mathrm{Imp}(X_i)=\mathbf{r}_i$ which consists of a ResNet18 \cite{he2016resnet} and an MLP head with a Sigmoid output layer, thus creating an $R$-dimensional score vector where each entry $r^k_i \in [0, 1]$ corresponds to the importance score for the relationship index $k \in \{1, ..., R\}$.
We train as follows: For each relationship index $k$ in a comparison pair $i$, let $r^{k,(0)}_i, r^ {k, (1)} _ i$ denote the predicted scores for each frame sequence. We define the score difference as
$
\delta^k_i = r^{k,(1)}_i -  r_i^{k, (0)} .
$
The strict-preference and tie sets are
$$
\mathcal{S}_{\mathrm{imp}} = \{i,k \mid y^k_i \neq 0\},
\qquad
\mathcal{M}_{\mathrm{imp}} = \{i,k \mid y^k_i = 0\}.
$$
and the strict preference loss is
\begin{equation*}
\mathcal{L}_{\mathrm{strict}} = \frac{1}{|\mathcal{S}_{\mathrm{imp}}|} \sum_{i,k \in \mathcal{S}_{\mathrm{imp}}} \log \left(1 + \exp\left(-\frac{y_i^k\delta^k_i}{\tau_{\mathrm{imp}}}\right)\right),
\end{equation*}
with the temperature $\tau_{\mathrm{imp}}$ as a hyperparameter. Additionally, we define a tie loss for comparisons in the tie set as
$$
\mathcal{L}_{\mathrm{tie}}
=
\frac{1}{|\mathcal{M}_{\mathrm{imp}}|}
\sum_{i,k \in \mathcal{M}_{\mathrm{imp}}}
\left[
\max(0, |\delta^k_i| - m_{\mathrm{tie}})
\right]^2,
$$
where the hyperparameter $m_{\mathrm{tie}}$ denotes the acceptable score difference margin. The total loss is
$
\mathcal{L}
=
\mathcal{L}_{\mathrm{strict}}
+
\lambda_{\mathrm{tie}} \mathcal{L}_{\mathrm{tie}},
$
with the weight hyperparameter $\lambda_{\mathrm{tie}}$.
After training, we score the relationship importances for each frame stack in each episode and assign the frame-wise labels via a moving average window.

\subsection{Task Progress Estimation}
In a similar fashion, we train a task progress labeler $\mathrm{Prog}(I_{1:T})=\mathbf{p}$, where $I_{1:T}$ is the complete frame sequence of task demonstration and $\mathbf{p}$ is a $T$-dimensional vector assigning each frame a task progress label. The task progress labeler consists of a ResNet18 that creates latent features $h_t = \mathrm{ResNet18}(I_t)$ from each frame in a task demonstration. The sequence of frame features $h_{1:T}$ is fed into a BiLSTM \cite{graves2005bilstm} which creates temporal features $u_{1:T} = \mathrm{BiLSTM}(h_{1:T})$. Finally, an MLP head predicts positive task progress increments
$$\Delta_t = \log(1+\mathrm{softplus}(\mathrm{MLP}(u_t))) + \epsilon,\qquad t=1,..., T-1.$$
We calculate a normalized temporal weight as
$$
w_t
=
\frac{\Delta_t}
{\max\left(\sum_{n=1}^{T-1} \Delta_n,\varepsilon\right)},
\qquad t = 1,\dots,T-1,
$$
from which we calculate the frame-wise progress as:
$$
p_1 = 0, \qquad p_t = \sum_{n=1}^{t-1}w_n,\qquad t=2,...,T.
$$
This calculation ensures continuously increasing task progress labels, ranging from 0 to 1, independent of the actual number of frames in a task demonstration. Since the monotonic construction already satisfies many strict progress comparisons, we filter the VLM comparison set to focus training on informative examples. The frame-wise progress predictions are first averaged over the same temporal windows of length $L=4$ used for the VLM comparisons, yielding one predicted progress value $\hat p_i^{(j)}$ per sequence. We define the predicted progress difference as
$
\delta_i^{\mathrm{prog}}
=
\hat p_i^{(0)}-\hat p_i^{(1)}.
$
We then filter equal-progress comparisons for consistency across demonstrations: comparisons whose temporal ordering conflicts strongly with other equal-progress comparisons are discarded, and only the most consistent fraction is retained. Since strict comparisons are often already satisfied by monotonicity, we preferentially retain those located temporally close to the retained equal-progress comparisons and randomly sample additional strict comparisons until a predefined strict-to-equal ratio is reached. The resulting filtered equal- and strict-progress comparison sets are denoted by $\widetilde{\mathcal{M}}_{\mathrm{prog}}$ and $\widetilde{\mathcal{S}}_{\mathrm{prog}}$, respectively
$\widetilde{\mathcal{S}}_{\mathrm{prog}}\subseteq\mathcal{S}_{\mathrm{prog}}$, $\widetilde{\mathcal{M}}_{\mathrm{prog}} \subseteq \mathcal{M}_{\mathrm{prog}}$ with
$$
\mathcal{S}_{\mathrm{prog}}=\{i\mid y_i^{\mathrm{prog}}\neq0\}, \quad \mathcal{M}_{\mathrm{prog}} = \{i | y_i^{\mathrm{prog}}=0\}.
$$
For the selected strict comparisons, we use a ranking loss
$$
\mathcal{L}_{\mathrm{rank}}
=
\frac{1}{|\widetilde{\mathcal{S}}_{\mathrm{prog}}|}
\sum_{i\in\widetilde{\mathcal{S}}_{\mathrm{prog}}}
\log
\left(
1+
\exp\left(
\frac{y_i^{\mathrm{prog}} \delta_i^{\mathrm{prog}}}{\tau_{\mathrm{prog}}}
\right)
\right),
$$
where \(\tau_{\mathrm{prog}}\) is a temperature hyperparameter. For retained equal-progress comparisons, we penalize differences exceeding a margin \(m_{\mathrm{same}}\),
$$
\mathcal{L}_{\mathrm{same}}
=
\frac{1}{|\widetilde{\mathcal{M}}_{\mathrm{prog}}|}
\sum_{i\in\widetilde{\mathcal{M}}_{\mathrm{prog}}}
\left[
\max
\left(
0,
|\delta_i^{\mathrm{prog}}|-m_{\mathrm{same}}
\right)
\right]^2.
$$
Additionally, $\mathcal{L}_{\mathrm{jump}}$ quadratically penalizes progress increments $w_t$ exceeding $\gamma/(T-1)$, discouraging the predicted progress from being concentrated in a small number of frames.
The complete training objective is
$$
\mathcal{L}
=
\lambda_{\mathrm{rank}}\mathcal{L}_{\mathrm{rank}}
+
\lambda_{\mathrm{same}}\mathcal{L}_{\mathrm{same}}
+
\lambda_{\mathrm{jump}}\mathcal{L}_{\mathrm{jump}}.
$$
After training, the progress labeler is applied to each complete task demonstration to obtain a continuous frame-wise task progress label \(p_t\in[0,1]\), which can subsequently be used to identify corresponding task stages across demonstrations.

\subsection{Relationship Representation Learning}
We represent each relationship by encoding the point cloud of the respective object pair into a low-dimensional latent vector $\mathbf{z}$. The point clouds of the object pairs are obtained by detecting and segmenting the task relevant objects as defined by the task description in a single frame from a task demonstration recording using GroundedSAM2 \cite{ren2024groundedsam}. Those initially segmented objects are then tracked through the whole task demonstration using SAM3 \cite{carion2025sam3}, producing fine-grained segmentation masks for each object.\\
The encoder is a DGCNN \cite{wang2019dgcnn} which is trained using a contrastive loss directly on the latent vector. This loss is an importance-weighted ranking loss which compares the latent vectors of an anchor sample $\mathbf{z}^a$ to latent vectors of one positive example $\mathbf{z}^+$ and three negative examples $\mathbf{z}^-_j, j =1, 2,3$. The final loss function is
\begin{equation*}
     \mathcal{L}_{\mathrm{rel}} = \frac{1}{3}\sum_{j=1}^{3}\max(0, \alpha^+d(\mathbf{z}^a, \mathbf{z}^+) - \alpha_{j}^-d(\mathbf{z}^a, \mathbf{z}_{j}^-) + m_{\mathrm{rel}}),
\end{equation*}
with a distance measure $d$ which we implement as the cosine distance $d(\mathbf{z}_1, \mathbf{z}_2) = 1 - \mathbf{z}_1^\top\mathbf{z}_2$ and a margin hyperparameter $m_{\mathrm{rel}}$. The triplet terms are weighted using the estimated relationship importance scores of each sample
\begin{equation*}
    \alpha^+ = \max(r^a,r^+), \qquad \alpha^-_j=\max(r^a, r^-_j),
\end{equation*}
with $r^a, r^+, r^-_j$ denoting the importance score of the anchor, positive sample, and negative samples. Intuitively, the loss function encourages decreasing the distance of the anchor latent $\mathbf{z}^a$ to the positive sample $\mathbf{z}^+$ while increasing the distance to the negative samples $\mathbf{z}^-_j, j =1, 2,3$. We collect the positive and negative samples by comparing task progress label, demonstration, and relationship identity to the anchor sample. The positive sample shows the same relationship, is taken from a frame with the same task progress label, but from a different demonstration. The negative samples either show a different relationship from the same demonstration at the same task progress, the same relationship at a different task progress within the same demonstration, or the same relationship at a different task progress in a different demonstration. Table \ref{tab:contrastive_samples} provides a compact overview of the positive and negative samples. Instead of considering the full combinatorial set of possible negative samples, we restrict training to three representative cases that capture distinct types of relationship mismatch. These cases are chosen to provide comparatively difficult negatives, under the assumption that relationships involving different object pairs are generally easier to distinguish than different states of the same object pair.

\begin{table}[]
    \centering
    \begin{tabular}{cccc}
        \hline\noalign{\vskip 1pt}
        \textbf{Task Progress} & \textbf{Demonstration} & \textbf{Relationship} & \textbf{Target}  \\
        \hline
        same & different & same & positive\\
        same & same & different & negative\\
        different & same & same & negative \\
        different & different & same & negative \\
        \hline
    \end{tabular}
    \caption{Overview of positive and negative samples for the relationship encoder contrastive loss.}
    \label{tab:contrastive_samples}
\end{table}

\subsection{Relationship-Aware Failure Prediction}
To detect erroneous relationship states, we apply OOD detection to the learned relationship latent $\mathbf{z}$. As detailed in section \ref{sec:related_work}, several methods can be trained using only successful in-distribution samples \cite{burda2019rnd, xu2025faildetect, liu2024pcakmeans}. We therefore do not introduce a new OOD detector. Instead, our relationship encoder provides a latent representation to which such detectors can be applied. While this makes our approach compatible with a broad range of OOD detectors, we choose RND \cite{burda2019rnd} due to its fast inference and comparatively high performance. We use FiLM \cite{perez2018film} to condition the RND OOD detector on the current task progress $p_t$.

For each relationship $k$, we train an independent detector $\phi_k$ on successful demonstrations and obtain an OOD score
$
s_t^k = \phi_k(\mathbf{z}_t^k, p_t).
$
A relationship-specific threshold $\theta_{\mathrm{ood}}^k$ is calibrated using CP \cite{angelopoulos2023conformalprediction}. To account for whether an anomalous relationship is relevant at the current task stage, we combine its OOD score with the predicted relationship importance $\hat r_t^k$. We use the importance as a gate,
\begin{equation*}
f_t^k =
\mathbb{I}[\hat r_t^k > \theta_{\mathrm{imp}}]
\mathbb{I}[s_t^k > \theta_{\mathrm{ood}}^k],
\quad f_t^{\mathrm{fail}} = \max_kf^k_t.
\end{equation*}
The importance threshold is a hyperparameter which we set to $\theta_{\mathrm{imp}} = 0.5$. An execution is classified as failed if $f_t^{\mathrm{fail}} = 1$ at any timestep.
During deployment, we use Cutie\cite{cheng2024Putting}, initialized with GroundedSAM2~\cite{ren2024groundedsam}, for object segmentation as it offers improved real-time tracking capabilities compared to SAM3.

\section{Experiments}
\label{sec:experiments}

\subsection{Experimental Setup}

\paragraph{Tasks}

\begin{figure}
    \centering
    \includegraphics[width=0.95\linewidth]{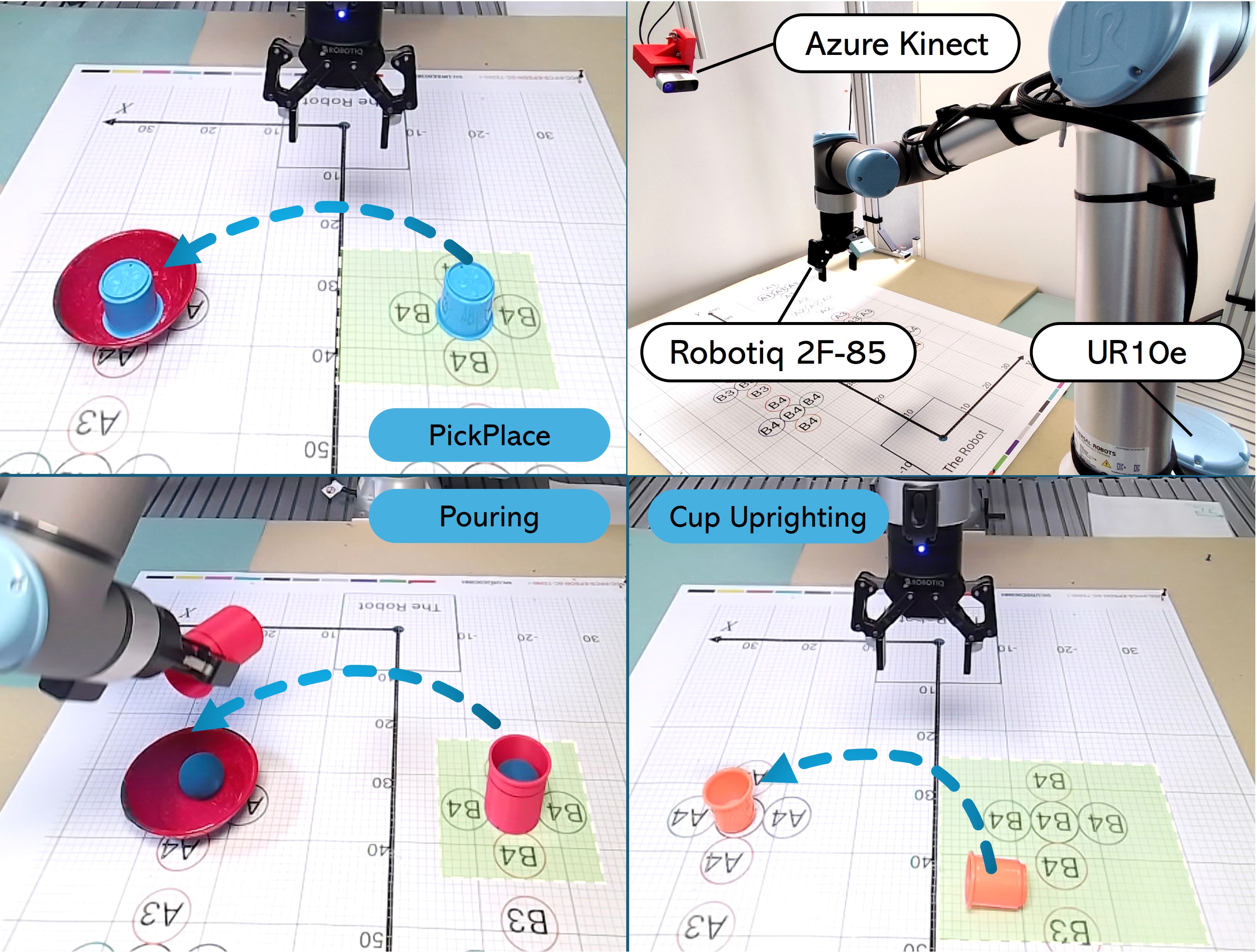}
    \caption{Experimental setup and representative real-world tasks used in the experiments. Marked regions indicate the variance in task setup.}
    \label{fig:tasks}
\end{figure}

We evaluate our method on three different manipulation tasks as visualized in Figure~\ref{fig:tasks}: \textit{PickPlace}, \textit{Uprighting}, and \textit{Pouring}, which require accurate grasping, insertion, and pouring interactions, respectively. In the \textit{PickPlace} task, the robot needs to pick a cylindrical object and place it inside a small bowl.
The \textit{Uprighting} task requires the robot to lift and upright a fallen cup.
In the \textit{Pouring} task, the robot needs to pick a cup and pour its content, a ball, into a small bowl.

\paragraph{Hardware and Data collection}
We record demonstrations via teleoperation using a Meta Quest 3 VR headset on a UR10e robot arm with a Robotiq 2F-85 parallel jaw gripper, which is also used during experiments. We use an Cartesian impedance controller\footnote{
\url{https://github.com/edgarwelteKIT/ur_impedance_driver_ros2}}
to prevent a safety stop during physical contact. The image and point cloud are captured by an external Azure Kinect camera. Figure~\ref{fig:tasks} visualizes the experimental setup. We collect 12 demonstrations for each task. The recordings consists of end-effector pose, gripper position, scene RGB/depth image, and point cloud, while the recorded action is the absolute end-effector pose and gripper position. Data is recorded at 10 Hz.
The recorded successful teleoperated demonstrations are used only for training/calibration. None of the 178 autonomous policy executions used in Tables \ref{tab:results_overview}-\ref{tab:ablations} are included in relationship encoder, OOD-detector, auxiliary-labeler training, or CP calibration.

\paragraph{Policy and Experiments}
We train the robot manipulation policy ManiFlow \cite{yan2025maniflow}, a flow matching policy for imitation learning, on the successful task demonstrations. The visual input is the scene point cloud. We extend the original architecture with two learnable output tokens: one for relationship-importance scores and one for task progress. Each token is mapped to its respective output dimension by a small MLP with a sigmoid output activation, and both heads are trained jointly with the policy using mean-squared-error losses. In contrast to the policies used by \cite{roemer2025failureprediction, xu2025faildetect}, ManiFlow does not force the visual input through a global bottleneck, creating an immediately usable, low-dimensional latent observation vector for OOD detection. Instead, ManiFlow deliberately retains local information by encoding each point as a token, thus mapping the input point cloud directly to a high dimensional token space, which leads to increased performance \cite{yan2025maniflow}. Using ManiFlow, we record 60 task executions for the \textit{Uprighting} and \textit{Pouring} tasks, and 58 task executions for the \textit{PickPlace} task. We induce OOD situations by variations of the placement of objects and by placing task unrelated objects in the background. This results in 42 failed and 18 successful executions for the \textit{Uprighting} task, 48 failed and 12 successful executions for the \textit{Pouring} task, and 45 failed and 13 successful executions for the \textit{PickPlace} task.

\subsection{Baselines}
As baselines, we choose successful and common real-time capable OOD- and uncertainty-based failure detectors. We choose Action Chunk Entropy (ACE), introduced by \cite{roemer2025failureprediction}, as the uncertainty-based baseline and also compare against the combination with RND~\cite{burda2019rnd} termed FIPER \cite{roemer2025failureprediction}. Our OOD-detector baselines are lopO~\cite{xu2025faildetect}, logpZO~\cite{xu2025faildetect}, PCA-kmeans~\cite{liu2024pcakmeans}, and RND~\cite{burda2019rnd}. For all of those we compare obtaining the necessary global observation latent vector by mean-pooling or max-pooling across the observation tokens and calculate separate thresholds using CP~\cite{angelopoulos2023conformalprediction}.

\subsection{Failure Detection Performance}

\begin{figure*}
    \centering
    \includegraphics[width=0.8\linewidth]{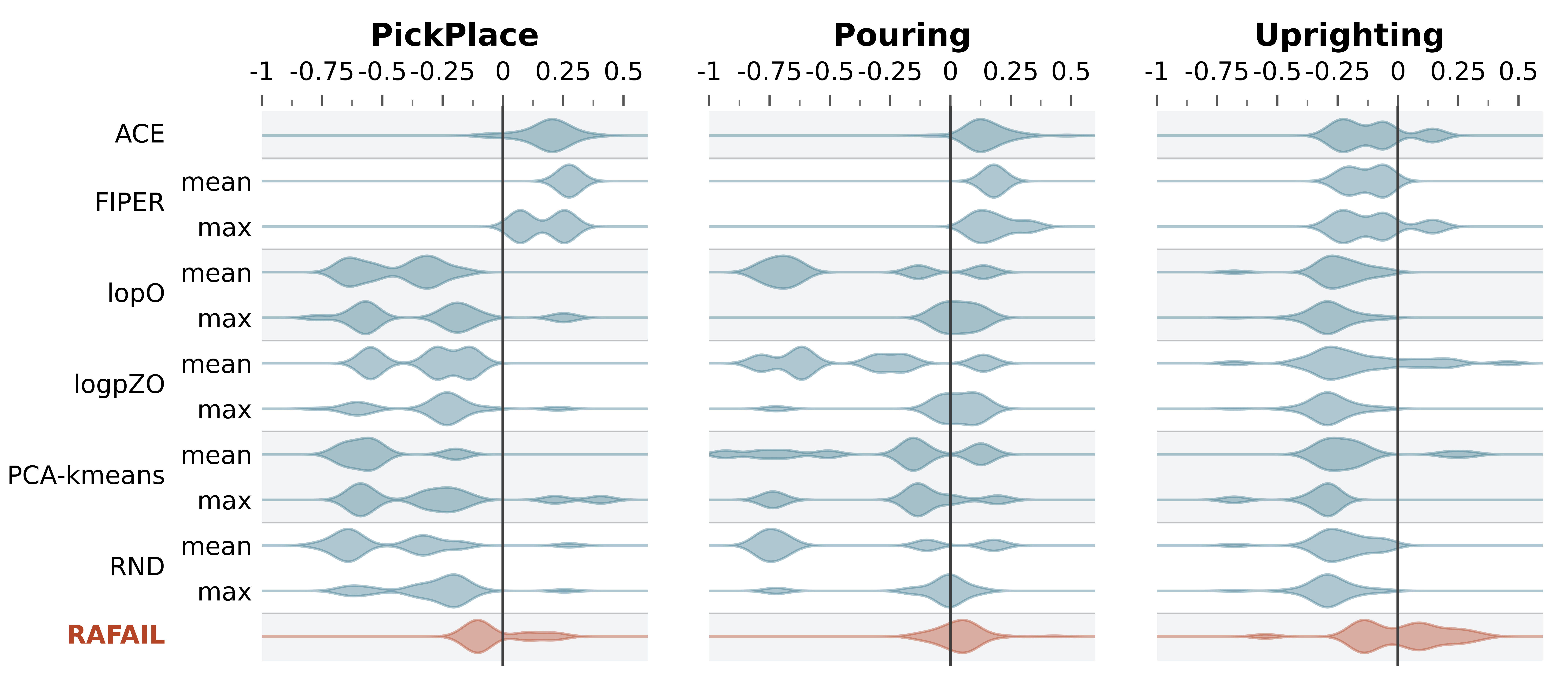}
    \caption{Temporal alignment of failure detections with observed failures. For each failed execution, we manually identify the timestep at which the failure occurs and use it as the reference timestep, corresponding to the vertical line at zero. The horizontal axis shows the difference between the first detection timestep and this reference, normalized by the maximum duration of successful executions. Negative values indicate detections before the failure and positive values detections after it. Each violin plot shows the distribution of true-positive detection times for one method and task. The "mean" and "max" variants refer to mean- or max-pooling as a dimensionality reduction.}
    \label{fig:temporal_distribution}
\end{figure*}

Table~\ref{tab:results_overview} compares the failure detection performance of our approach to the baselines for each task and across all evaluated executions (marked "Overall"). We report the balanced accuracy ($\mathrm{Acc.}$), the true positive rate ($\mathrm{TPR}$), and the false positive rate ($\mathrm{FPR}$) with
\begin{equation*}
    \text{Acc.} = \frac{1}{2}(\text{TPR} + 1 - \text{FPR}), \quad \text{TPR} = \frac{\text{TP}}{\text{P}}, \quad \text{FPR} = \frac{\text{FP}}{\text{N}},
\end{equation*}
where $\text{P}$ is the number of failed task executions (positives), $\text{N}$ is the number of successful task executions (negatives), $\text{TP}$ is the number of failed task executions that were correctly detected as failures (true positives), and $\text{FP}$ is the number of successful task executions that were wrongly detected as failures (false positives).

\begin{table*}[]
    \centering
    \begin{tabular}{llcccccccccccc}
    \hline\noalign{\vskip 2pt}
        Task & Metric & ACE \cite{roemer2025failureprediction} & \multicolumn{2}{c}{FIPER \cite{roemer2025failureprediction}} &\multicolumn{2}{c}{lopO \cite{xu2025faildetect}}  & \multicolumn{2}{c}{logpZO \cite{xu2025faildetect}} & \multicolumn{2}{c}{PCA-kmeans \cite{liu2024pcakmeans}} & \multicolumn{2}{c}{RND \cite{burda2019rnd}} & RAFAIL (ours) \\
         & & & mean & max & mean & max & mean & max & mean & max & mean & max & \\[2pt]
         \hline\hline\noalign{\vskip 2pt}
          & Acc. (\%) $\uparrow$  & 64.0 & 51.1 & 53.3 & 49.1 & 64.5 & 53.3 & \underline{68.0} & 52.8 & 52.9 & 56.8 & \textbf{72.8} & 66.7\\
         \textsc{PickPlace} & TPR (\%) $\uparrow$  & 51.1 & 2.2 & 6.7 & 28.9 & 44.4 & 6.7 & \textbf{82.2} & 13.3 & 28.9 & 44.4 & \underline{53.3} & 33.3\\
        & FPR (\%) $\downarrow$ & 23.1 & \textbf{0.0} & \textbf{0.0} & 30.8 & 15.4 & \textbf{0.0} & 46.2 & \underline{7.7} & 23.1 & 30.8 & \underline{7.7} & \textbf{0.0}\\
         \hline\noalign{\vskip 2pt}
         & Acc. (\%) $\uparrow$ & \underline{71.9} & 51.0 & 51.0 & 53.1 & 60.4 & 40.6 & 61.5 & 62.5 & 59.4 & 57.3 & 52.1 & \textbf{82.3}\\
         \textsc{Pouring} & TPR (\%) $\uparrow$ & 68.8 & 2.1 & 10.4 & 8.3 & 29.2 & 14.6 & 31.3 & 25.0 & 18.8 & 14.6 & 20.8 & \textbf{81.3}\\
         & FPR (\%) $\downarrow$ & 25.0 &  \textbf{0.0} & \underline{8.3} & 14.6 & \underline{8.3} & 33.3 & \underline{8.3} & \textbf{0.0} & \textbf{0.0} & \textbf{0.0} & 16.7 & 16.7\\
         \hline\noalign{\vskip 2pt}
         & Acc. (\%) $\uparrow$  & 55.7 & 54.8 & 54.4 & 61.1 & 50.0 & 63.1 & 50.0 & \underline{65.1} & 47.2 & 64.7 & 50.0 & \textbf{71.4}\\
         \textsc{Uprighting} & TPR (\%) $\uparrow$  & 16.7 & 9.5 & 14.3 & 50.0 & \textbf{100.0} & \underline{59.5} & \textbf{100.0} & 35.7 & 16.7 & 57.1 & \textbf{100.0} & \underline{59.5}\\
         & FPR (\%) $\downarrow$ & \underline{5.6} & \textbf{0.0} & \underline{5.6} & 27.8 & 100.0 & 33.3 & 100.0 & \underline{5.6} & 22.2 & 27.8 & 100.0 & 16.7\\
         \hline\noalign{\vskip 2pt}
         & Acc. (\%) $\uparrow$  & \underline{65.2} & 52.2 & 52.9 & 53.6 & 53.8 & 51.3 & 55.7 & 59.9 & 52.6 & 58.4 & 53.7 & \textbf{73.4}\\
         Overall & TPR (\%) $\uparrow$  & 46.7 & 4.4 & 10.4 & 30.4 & 56.3 & 26.0 & \textbf{69.6} & 24.4 & 21.5 & 37.8 & 56.3 & \underline{58.5}\\
         & FPR (\%) $\downarrow$ & 16.3 & \textbf{0.0} & \underline{4.7} & 23.3 & 48.8 & 23.3 & 58.1 & \underline{4.7} & 16.3 & 20.9 & 48.8 & 11.6\\
         \hline
    \end{tabular}
    \caption{Failure detection performance across real-world robotic manipulation tasks. We report balanced accuracy (Acc.), true positive rate (TPR), and false positive rate (FPR) in percent. The "mean" and "max" variants refer to mean- or max-pooling as a dimensionality reduction. Best results are marked in bold and second best by underlining.}
    \label{tab:results_overview}
\end{table*}

Across all task executions, RAFAIL achieves the highest Acc. of 73.4\% by combining a comparatively high TPR of 58.5\% with a relatively low FPR of 11.6\%. The second best overall method is ACE~\cite{roemer2025failureprediction} which achieves an Acc. of 65.2\%, a lower TPR of 46.7\% and a higher FPR of 16.3\%. RAFAIL achieves the highest Acc. for the \textit{Pouring} task, where it also achieves the highest TPR, and for the \textit{Uprighting} task. For the \textit{PickPlace} task, RND and logpZO with max-pooling achieve the highest Acc. of 72.8\% and 68.0\% while RAFAIL reports the third-highest Acc. of 66.7\%. It can be observed that several baseline methods struggle to maintain a consistent trade-off between TPR and FPR across tasks. Depending on the task, some methods appear to become either overly sensitive, with high TPR and FPR, or overly conservative, with low FPR but also low TPR. This suggests sensitivity to task-specific failure signals and limited robustness of a single detection configuration across tasks. Variants that use max-pooling appear to be overly sensitive, while mean pooling appears to be somewhat more stable, although often at the cost of lower TPR. Likely, this is because of the smoothing effect of mean-pooling, while max-pooling emphasizes the strongest response, increasing sensitivity but also susceptibility to benign variations. ACE shows comparatively consistent behavior, while RAFAIL maintains the most favorable TPR/FPR trade-off across all tasks, resulting in consistently high balanced accuracy.

We manually annotate the timestep in which the failure occurs in each task execution and record the timestep when failure is detected for each detection method. This manually annotated failure timestep serves as a reference rather than an independent ground truth, as the exact timestep of a failure can be ambiguous and may be annotated slightly differently by another observer. We compute the normalized signed detection delay as the difference between the detection timestep, divided by the maximum total execution time of successful executions, with negative values indicating early detection and positive values indicating late detection. The violin plots in Figure~\ref{fig:temporal_distribution} show smoothed distributions. RAFAIL tends to trigger closer to the annotated failure timestep, especially for \textit{PickPlace} and \textit{Pouring}. Several baselines instead show distributions shifted substantially before or after the failure, suggesting that their OOD signals may respond to deviations that are less directly associated with the observed task failure. For \textit{Uprighting}, RAFAIL shows a broader distribution. This may partly result from the nature of the task, where failures can develop gradually and may expose detectable relationship deviations at different times relative to the annotated failure timestep. Overall, the results in Figure~\ref{fig:temporal_distribution} suggest that the OOD signals produced by RAFAIL are more closely aligned with the physical manifestation of failure than those produced by the baselines. This may partially explain RAFAIL's more consistent TPR/FPR trade-off across tasks, as its detections tend to occur closer to the observed failure event.

To further assess whether the focus on individual relationships provides meaningful information about the failure, we manually annotate the relationships involved in each observed failure and compare them with the relationship triggering the failure in RAFAIL. Among true-positive detections, the agreement between the relationship triggering the failure detection and the annotated relationship is 93.3\% for the \textit{PickPlace} task, 69.3\% for the \textit{Pouring} task, and 68.0\% for the \textit{Uprighting} task.
While these annotations can be subjective and therefore are not an independent ground truth, the high agreement suggests that RAFAIL frequently identifies relationships that are consistent with the observed failure.

\subsection{Ablations}
We compare encoding and supervising individual relationships to encoding the whole scene by training and applying our relationship encoder to the whole scene point cloud. During the training, we use scene point clouds at the same task progress as positive and at different task progress as negative samples. While the relationship importance is not used in this setting, we still condition the OOD detection on the predicted task progress. Additionally, we investigate the role of the predicted relationship importance by comparing to a version of RAFAIL that does not take the relationship importance into account during OOD detection which can easily be achieved by setting $\theta_{\mathrm{imp}}=0$ which removes the importance gate and considers all relationships to be important all the time. Additionally, we investigate using the predicted relationship importance to continuously weight the OOD scores using
$
f_t^k = \mathbb{I}[\hat r^k_ts_t^k > \theta_{\mathrm{ood}}^k],
$
rather than a gate.
Table~\ref{tab:ablations} shows the resulting Acc., TPR, and FPR across the three tasks. It indicates that both the relationship-wise encoding and the predicted relationship importance add to the final performance. Encoding the whole scene results in lower balanced accuracy across all tasks, suggesting that focusing the observation on individual relationships provides more suitable failure signals. Removing the influence of the predicted relationship importance by setting $\theta_{\mathrm{imp}}=0$ increases the TPR, but also the overall FPR from 11.6\% to 69.8\%, reducing the Acc. from 73.4\% to 58.1\%. Because the gate can only suppress detections, this trade-off can be task dependent: the $\theta_{\mathrm{imp}}=0$ variant maintains a consistently high TPR and achieves the highest Acc. for \textit{Pouring}, where the relationship-wise OOD scores appear to separate successful and failed executions sufficiently well, but its high FPR makes it ineffective for \textit{PickPlace} and \textit{Uprighting}. This suggests that the predicted relationship importance provides a useful signal for suppressing OOD detection in currently less relevant relationships. Using the importance as a continuous weight rather than a gate results in a more conservative detector, with an improved overall FPR but also a reduced TPR of 37.8\%. Its overall Acc. of 68.9\% is however above that of the best evaluated baseline, indicating that both formulations can be effective while the gate provides the more favorable trade-off in the evaluated tasks.

\section{Conclusion}
We introduced RAFAIL, a relationship-aware framework for runtime failure detection that learns task-relevant object relationships from successful demonstrations. RAFAIL avoids runtime VLM inference; online semantic processing is limited to object tracking, relationship encoding, and RND evaluation. Across three real-world manipulation tasks, RAFAIL achieved 73.4\% balanced accuracy, outperforming the strongest evaluated baseline while maintaining a comparatively low false-positive rate.  The relationship-specific detections also frequently agreed with the relationships involved in the observed failures, supporting the usefulness of the relational representation.
However, subtle failures caused by small pose deviations may remain within the distribution of successful relationship states and therefore go undetected. Future work should investigate finer-grained relational representations for detecting such errors.
RAFAIL currently assumes that task-critical failure modes can be represented by pairwise relations among segmentable entities identified from successful demonstrations. Failures involving unmodeled entities, global scene properties, contact forces not visible geometrically, or segmentation/tracking breakdown may therefore remain undetected.

\begin{table}[]
    \centering
    \begin{tabular}{llcccc}
    \hline\noalign{\vskip 2pt}
        Task & Metric & \multicolumn{3}{c}{RAFAIL} & Scene \\
         & & gate & weight & $\theta_{\mathrm{imp}}=0$ &  \\[2pt]
         \hline\hline\noalign{\vskip 2pt}
          & Acc. (\%) $\uparrow$  & \textbf{66.7} & \underline{63.3} & 44.9 & 23.2\\
         \textsc{PickPlace} & TPR (\%) $\uparrow$  & \underline{33.3} & 26.7 & \textbf{66.7} & 31.1\\
        & FPR (\%) $\downarrow$ & \textbf{0.0} & \textbf{0.0} & \underline{76.9} & 84.6\\
         \hline\noalign{\vskip 2pt}
         & Acc. (\%) $\uparrow$ & 82.3 & \underline{86.5} & \textbf{87.5} & 38.5\\
         \textsc{Pouring} & TPR (\%) $\uparrow$ & \underline{81.3} & 72.9 & \textbf{91.7} & 77.1\\
         & FPR (\%) $\downarrow$ & \underline{16.7} & \textbf{0.0} & \underline{16.7} & 100.0\\
         \hline\noalign{\vskip 2pt}
         & Acc. (\%) $\uparrow$  & \textbf{71.4} & \underline{54.8} & 50.0 & \underline{54.8}\\
         \textsc{Uprighting} & TPR (\%) $\uparrow$  & \underline{59.5} & 9.5 & \textbf{100.0} & 9.5\\
         & FPR (\%) $\downarrow$ & \underline{16.7} & \textbf{0.0} & 100.0 & \textbf{0.0}\\
         \hline\noalign{\vskip 2pt}
         & Acc. (\%) $\uparrow$  & \textbf{73.4} & \underline{68.9} & 58.1 & 43.6\\
         Overall & TPR (\%) $\uparrow$  & \underline{58.5} & 37.8 & \textbf{85.9} & 40.7\\
         & FPR (\%) $\downarrow$ & \underline{11.6} & \textbf{0.0} & 69.8 & 53.5\\
         \hline
    \end{tabular}
    \caption{Ablations of failure detection performance of our proposed RAFAIL (RAFAIL gate) and variants that use relationship importance as a continuous weight (RAFAIL weight), ignore relationship importance (RAFAIL $\theta_{\mathrm{imp}}=0$), or encode the whole scene instead of individual relationships (Scene).}
    \label{tab:ablations}
\end{table}

\bibliographystyle{IEEEtran}
\bibliography{bibliography}

\end{document}